\documentclass[letterpaper, 10 pt, conference]{ieeeconf}
\IEEEoverridecommandlockouts

\usepackage{url}
\usepackage{orcidlink}
\usepackage{hyperref}
\hypersetup{
    colorlinks=true,
    linkcolor=blue,
    filecolor=magenta,      
    urlcolor=black,
    pdftitle={A Multi-Modal Perception Pipeline for Object Detection and Tracking in Autonomous Racing},
    pdfpagemode=FullScreen,
    }
\usepackage{graphicx}
\usepackage{cite}
\usepackage{amsmath,amssymb,amsfonts}
\usepackage{algorithmic}
\usepackage{graphicx}
\usepackage{textcomp}
\usepackage{xcolor}
\usepackage{svg}
\usepackage{censor}
\usepackage{booktabs}
\usepackage{amssymb}
\usepackage{pifont}
\newcommand{\cmark}{\ding{51}}%
\newcommand{\xmark}{\ding{55}}%
\StopCensoring
\usepackage[dvipsnames]{xcolor}
\def\BibTeX{{\rm B\kern-.05em{\sc i\kern-.025em b}\kern-.08em
    T\kern-.1667em\lower.7ex\hbox{E}\kern-.125emX}}
\makeatletter 
\newcommand{\linebreakand}{%
  \end{@IEEEauthorhalign}
  \hfill\mbox{}\par\vspace{-0.2cm}
  \mbox{}\hfill\begin{@IEEEauthorhalign}
}
\makeatother 

\title{\LARGE \bf A Multi-Modal Perception Pipeline for \\Object Detection and Tracking in Autonomous Racing
}

\author{
	\parbox{\textwidth}{%
		\centering
        \hspace{-0.21cm}
		Davide Malvezzi$^{1}$, Michele Pestarino$^{1}$, Vittoria Cavicchioli$^{2}$, Valentina La Gamba$^{2}$,\\ Silvia Severi$^{2}$, Fabio Bagni$^{2}$, Luca Bartoli$^{2}$, Massimiliano Bosi$^{2}$, Francesco Gatti$^{2}$,\\ Micaela Verucchi$^{2}$, Ayoub Raji$^{1}$, Marko Bertogna$^{1,2}$
	}%
	\thanks{$^{1}$University of Modena and Reggio Emilia, Modena, Italy
		{\tt\small $\{$davide.malvezzi, michele.pestariono, ayoub.raji, marko.bertogna$\}$@unimore.it}}%
	\thanks{$^{2}$HiPeRT Srl, Modena, Italy
		{\tt\small 
        $\{$vittoria.cavicchioli,
        valentina.lagamba, 
        silvia.severi, 
        fabio.bagni, 
        luca.bartoli, 
        massimiliano.bosi, 
        francesco.gatti, 
        micaela.verucchi, 
        marko.bertogna$\}$@hipert.it
        }}%
}

\newcommand\copyrighttext{%
\footnotesize \copyright 2026 IEEE. Personal use of this material is permitted. Permission from IEEE must be obtained for all other uses, in any current or future media, including reprinting/republishing this material for advertising or promotional purposes, creating new collective works, for resale or
redistribution to servers or lists, or reuse of any copyrighted component of this work in other works.}
\newcommand\copyrightnotice{%
\begin{tikzpicture}[remember picture,overlay]
\node[anchor=north,yshift=-1cm] at (current page.north) {\parbox{\dimexpr\textwidth-\fboxsep-\fboxrule\relax}{\centering \copyrighttext}};
\end{tikzpicture}%
}

\begin{document}

\maketitle
\copyrightnotice
\thispagestyle{empty}
\pagestyle{empty}

\begin{abstract}
Object detection and tracking are fundamental components of perception systems for autonomous driving. Achieving robust performance under adverse conditions such as limited visibility, sensor noise, and failures remains an open challenge, particularly in autonomous racing, where vehicles operate at very high speeds, experience strong vibrations, and interact under small safety margins. This paper presents a multi-modal late-fusion perception pipeline for object detection and tracking in the autonomous racing domain. The proposed system extends previous work by exploiting all onboard sensors through a late-fusion approach and a dedicated multi-object tracking framework. Independent detections from cameras, LiDARs, and RADARs are combined to provide timely and robust state estimates of surrounding vehicles. The tracking method explicitly compensates for detection delays and embeds in its model prior knowledge of vehicle dynamics and track layout.
Experimental evaluation on real-world data across diverse critical scenarios, representative of challenging edge cases also in urban driving, confirms the effectiveness of the proposed pipeline and its suitability to support safe and adaptive planning decisions.
\end{abstract}


\section{INTRODUCTION}
\label{introduction}
Autonomous racing represents one of the most extreme operational domains for perception systems. At speeds approaching 80 m/s, even small delays can translate into meter-scale state estimation errors: a latency of 100 ms corresponds to nearly 8 meters of spatial displacement. In such conditions, asynchronous multi-sensor measurements, detection delays, and high relative velocities significantly amplify tracking inaccuracies and directly impact collision avoidance and overtaking safety. The latest research trends focus on reducing inference time, improving Bird-Eyed-View (BEV) representations and operators' performance, and achieving more accurate depth estimation from cameras \cite{Mao_2023}.
Another important goal in this domain is to achieve robust performance under adverse weather conditions, limited visibility, and sensor failures. These are scenarios highly relevant in the motor-sport domain, and in this regard, the increase in research on Autonomous Racing can be exploited as an experimental and learning field for autonomous driving\cite{betz}. In particular, the birth of recent international competitions, such as the Indy Autonomous Challenge (IAC) in 2021 and the Abu Dhabi Autonomous Racing League (A2RL) in 2024, helped to deploy and test the state-of-the-art solutions in full-scale autonomous race cars running at speeds up to 80 m/s, with a high differential velocity between different agents, high vibration and noise that stress the sensors, and in conditions where latency and delay are extremely important.
\begin{figure}[tb]
\centering
{\includegraphics[width=\linewidth]{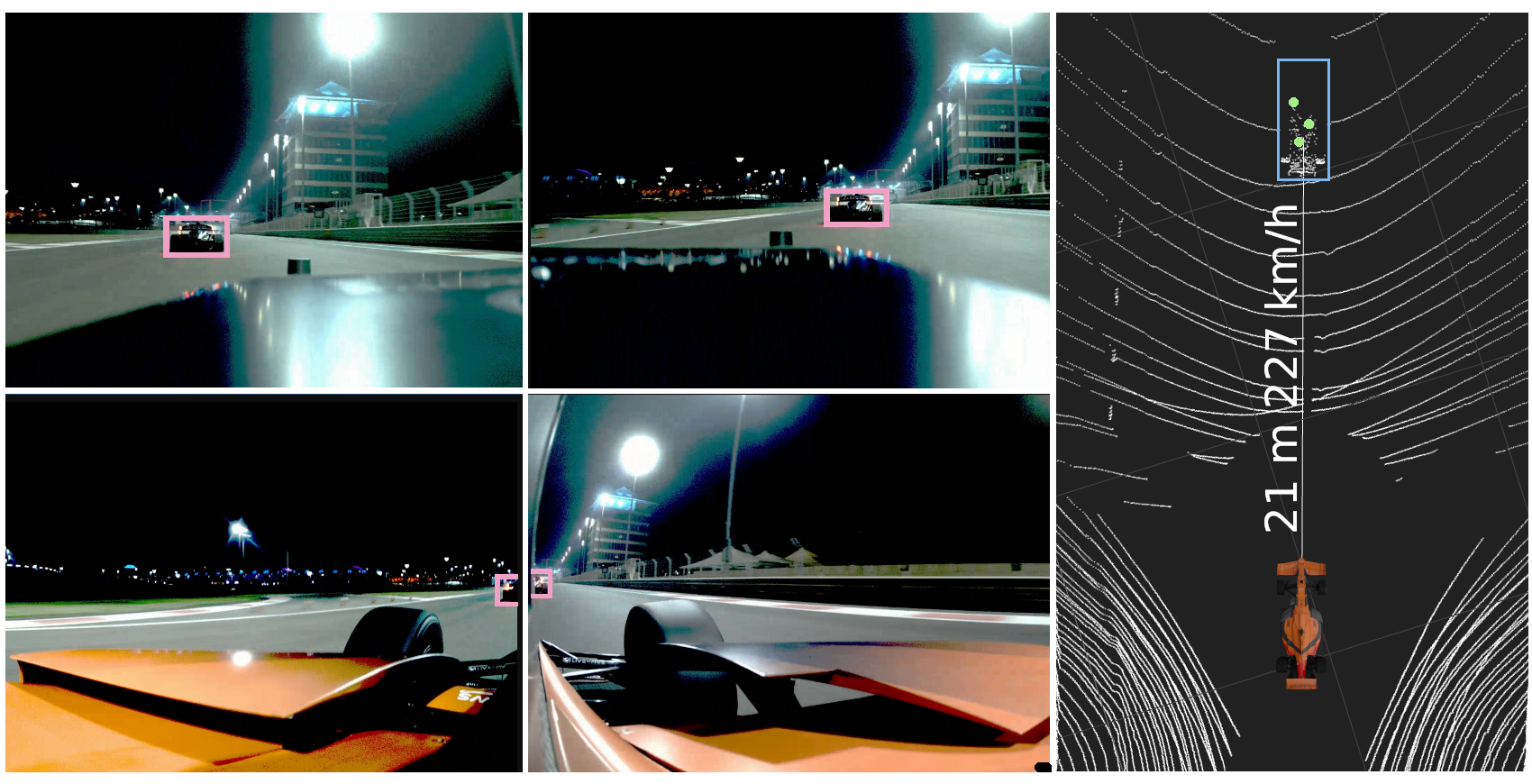}}
\caption{UNIMORE Racing's Dallara EAV-24 competing in the 2025 A2RL at the Yas Marina F1 Circuit. Left: detections from four cameras (\textcolor{pink}{pink}). Right: Bird’s-Eye View showing LiDAR detection (\textcolor{blue}{blue}) and RADAR detections (\textcolor{Green}{green}), together with the estimated distance and velocity of the opponent.}
\label{fig:detections}
\end{figure}
In this paper, we present a perception pipeline for object detection and tracking in the autonomous racing domain, deployed by \censor{the UNIMORE Racing team} during the 2025 A2RL event at the Yas Marina Circuit (Fig. \ref{fig:detections}). The proposed solution 
takes advantage of all sensors onboard through a late fusion approach and an ad-hoc object tracking process that explicitly accounts for delay compensation. 
Section~\ref{related_works} provides an extensive overview of the state-of-the-art in perception-related research, highlighting the existing gaps in the literature that this work aims to address. Section~\ref{system_overview} describes the hardware platform of the experimental vehicle and presents an overview of the proposed perception pipeline. The detection processes adopted for each sensor modality are detailed in Section~\ref{detections}, while Section~\ref{detections_fusion} explains how the detections of different sensors are fused together. The object tracking methodology is presented in Section~\ref{object_tracking}. Section~\ref{results} reports the experimental evaluation, discussing both the performance and limitations of the proposed approach in three different scenarios typical of racing competitions. Finally, Section~\ref{conclusion} concludes the paper and outlines directions for future work.

\section{RELATED WORK}
\label{related_works}
The literature on perception in autonomous racing can be broadly categorized into three categories: creation of multi-modal datasets, single-modality detection pipelines, and partial multi-sensor fusion approaches.
In \cite{Kulkarni_2023}, the first-ever open dataset for high-speed autonomous racing is presented with more than 6.5 hours of data in ROS2 and nuScenes format, with some benchmarks on object detection and tracking. In particular, the authors compared the average precision of LiDAR detections using BEV labels and 3D bounding boxes with PointPillars \cite{POINTPILLARS} and Voxel-R-CNN \cite{deng2020voxel}, achieving acceptable precision at a maximum range of 60 m. A brief presentation of RADAR-based detection and tracking is given, focusing on noise reduction and filtering, while for the cameras, a solution based on YOLOv5 is evaluated, with detections achieving distances over 100 meters. Despite the fact that all available sensors have been tested for the detection task, no sensor fusion or multi-modal detection approaches are presented. Another dataset is presented in \cite{ghosh2025racingdatasetbaselinemodel}, which focuses on providing annotated multi-camera images for detection on road courses, in particular, lane detection, proposing a Generative Adversarial Network (GAN) method and comparing it to current state-of-the-art solutions. 
The most recent dataset in the autonomous racing domain is BETTY~\cite{BETTY_DATASET}, a large-scale, multi-modal dataset collected from multiple autonomous racing vehicles. It targets a variety of autonomous driving tasks and comprises more than 13 hours of recorded data. However, no benchmark of the state-of-the-art method for object detection and tracking is provided on the presented data.
The most relevant works on object detection and tracking for autonomous racing are presented by researchers who took part in the IAC and A2RL events. Team PoliMOVE presented a solution based on LiDAR sensors in which a tracking-by-detecting approach is used, following classical algorithms for point cloud segmentation with ground removal and clustering, pose estimation exploiting knowledge of track layout, and object tracking via a variable-step Extended Kalman Filter (EKF)\cite{cellina2025}. In \cite{Jung_2023}, team KAIST described their solution based on LiDAR clustering, with ground filtering on the 3D point clouds via height distribution, fused and tracked, including the RADARs detections via an Interacting Multiple Model Filter employing three models: constant velocity, constant turn rate and velocity, and random motion. Their results showed the best accuracy and recall using LiDAR-only clustering and tracking in a detection range lower than 50 meters, while in a mid-range distance (50-100 m), exploiting both sensors brought better performance, and finally, at distances greater than 100 m only the RADAR-based detections were valuable with a low recall due to the limited range of the LiDAR sensors. Another work from the KAIST team presents MonoDINO-DETR, a monocular 3D object detection framework that employs a Vision Foundation Model (DINOv2) as the backbone to extract global features for depth estimation  \cite{KAIST_MONODINODETR}. The reported results demonstrate competitive performance. However, inference time may limit its applicability in strict real-time scenarios, and the relatively small size of the training and validation datasets suggests that further evaluation on larger benchmarks would be beneficial to fully assess its generalization capability.
The object tracking method presented by the TUM Autonomous Motorsport team in~\cite{TUM_PERCEPTION} is also based on the fusion of LiDAR and RADAR detections. Particular emphasis is placed on compensating for sensor-specific detection delays, as at high speeds, the apparent position of an object can shift by several meters if such delays are not accounted for. In addition, a plausibility check is applied to filter out detections that lie outside the drivable area of the track. Finally, Team MIT–PITT presented their autonomous racing stack in \cite{MIT_PIT_PERCEPTION}. Their perception module relies solely on cameras and LiDAR, with RADAR sensors not being utilized. YOLOv5 is used to detect race cars in camera images, and the 3D object pose is recovered by exploiting prior knowledge of the shape and size of the vehicles. However, the pose estimation error increases proportionally with the real-world distance between the camera and the target vehicle. For LiDAR data, PointPillars is used to perform object detection on point clouds. Object association and tracking are then carried out using AB3DMOT~\cite{AB3DMOT}. Notably, no multi-modal sensor fusion is applied prior to object association and tracking.
In these works, not all available sensor modalities have been exploited simultaneously, due to factors such as computational limitations, slow overall pipeline execution relative to the demands of high-speed racing, and elevated false positive rates. In addition, critical components such as delay compensation have been implemented only by team TUM, and none of the existing tracking methods incorporate racing-specific knowledge. To address these gaps, we propose a unified, delay-aware, multi-modal perception architecture that integrates cameras, LiDARs, and RADARs within a consistent tracking framework explicitly designed for extreme-speed racing scenarios. We validate the proposed approach through experimental evaluations that highlight both its performance and limitations in three scenarios representative of typical racing conditions.

\section{System Overview}
\label{system_overview}
\begin{figure}[b]
\centering
{\includegraphics[width=0.85\linewidth]{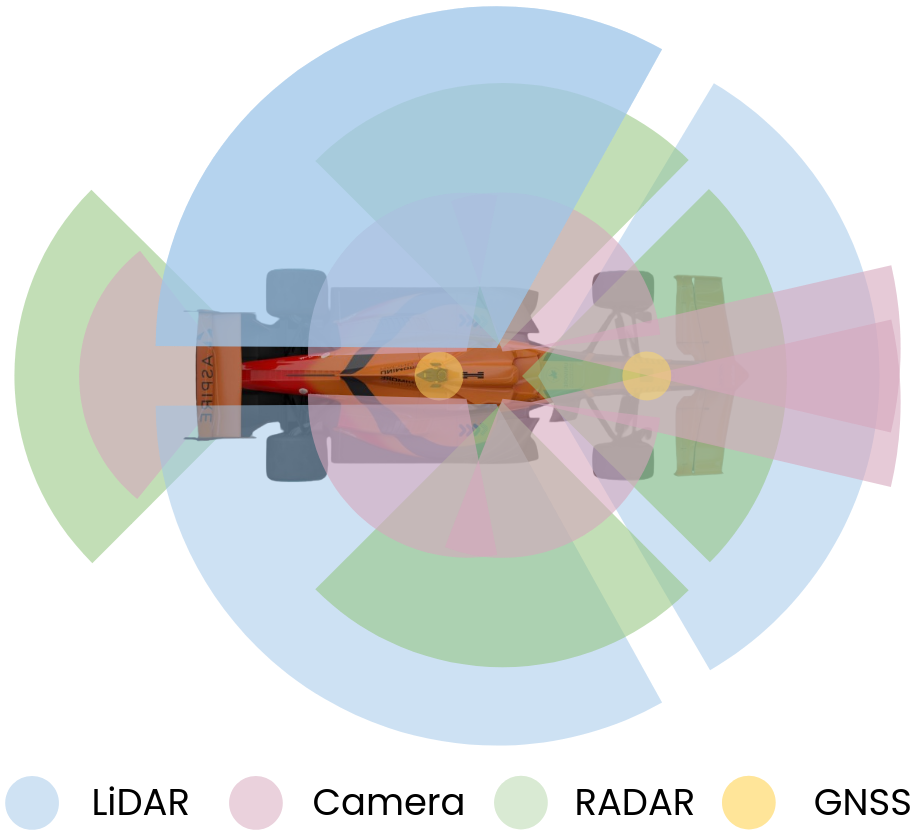}}
\caption{Sensor setup for the Dallara EAV-24.}
\label{fig:sensors}
\end{figure}

The Dallara EAV-24 is equipped with a comprehensive array of sensors, including a Vectornav VN-310 dual-antenna GNSS units and inertial measurement solutions. In addition to these, the vehicle features three Innovusion Falcon K LiDAR sensors, seven Sony IMX728 cameras, and four ZF ProWave RADAR units. The sensor setup on the vehicle is shown in Fig. \ref{fig:sensors}.

\begin{figure*}[tb]
\centering
\vspace{0.5cm}
{\includegraphics[width=1\textwidth]{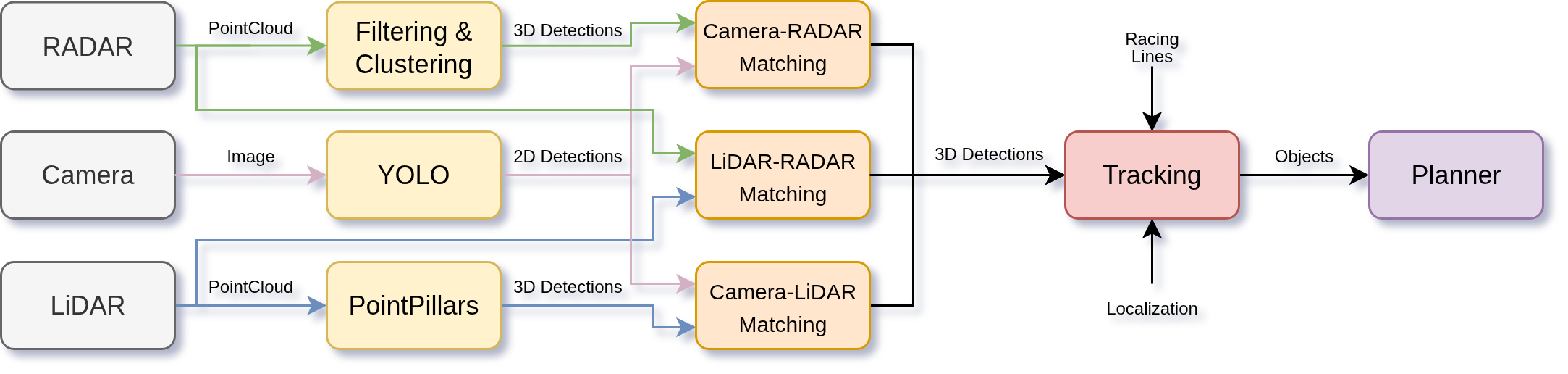}}
\caption{In the proposed detection and tracking pipeline, object detections are extracted for each sensor modality and subsequently fused together. The fused detections are used as measurement updates by a multi-object tracker, whose outputs are finally provided to the planning module to compute evasive maneuvers.}
\label{fig:pipeline}
\end{figure*}

The object tracking pipeline combines detections from individual sensors using a late fusion approach. Late fusion enables the integration of heterogeneous sensor modalities, providing increased flexibility. Changes can be introduced at the fusion stage without requiring a redesign of the entire system. Moreover, late fusion improves the robustness of the overall pipeline, as in the event of a sensor failure, the remaining sensors can continue to provide object detections, thereby ensuring continuity of the tracking.
However, improper temporal synchronization or spatial alignment of the data sources in the late fusion pipeline can lead to inconsistencies in the final detection results, causing incorrect detection associations. Therefore, all sensors are temporally synchronized using the Precision Time Protocol (PTP) \cite{PTP}, sharing the clock with the main computation system. Furthermore, an offline multi-sensor calibration procedure is carried out to accurately estimate the relative poses of all sensors mounted on the race car. First, the intrinsic parameters of the camera and the distortion coefficients of the lens are obtained using a checkerboard-based calibration method \cite{CAMERA_CALIBRATION}. Subsequently, the extrinsic pose between cameras and LiDARs with overlapping field-of-view is computed using an ArUco-based calibration approach \cite{CAMERA_LIDAR_CALIBRATION}. Finally, the position and orientation of the RADARs are estimated using a velocity-based calibration following the method proposed by \cite{RADAR_CALIBRATION}.

A schematic representation of the proposed object detection and tracking pipeline is shown in Fig. \ref{fig:pipeline}. The process begins by extracting object detections from each sensor modality, with a dedicated detection method applied to each. In Fig. \ref{fig:detections}, examples of the resulting detections are illustrated. Using the sensor calibration information, the fusion module associates detections from sensors with overlapping fields of view, ensuring that observations of the same physical object are correctly matched. The fused detections are then used to update a multi-object tracking module, which continuously estimates the position and velocity of each object. Finally, the tracked objects are provided as input to the planning module, which uses the estimated states to plan overtaking maneuvers \cite{OT_LOGIC}.

 
\section{Detections}
\label{detections}
\subsection{Camera Detections}
\label{camera_detections}
Camera-based object detection is performed using the YOLOv4 neural network \cite{YOLOV4}. The model is initially pre-trained on the open-source DeepDrive BDD100k dataset \cite{BBD100k} to recognize road objects such as cars, bikes, and pedestrians. The model is then specifically fine-tuned using a custom dataset, which includes data collected from online videos of human-driven open-wheel race cars and nearly 600 manually labeled images from previous test sessions.
During the inference process, the model receives a batch of \( N_{\text{cameras}} \) images and for each camera $j$ outputs a set of 2D detections (\(\textit{BB}^{\text{2D}}_{\text{cam}_j}\)). Each detection \(bb_{ji} \in \textit{BB}^{\text{2D}}_{\text{cam}_j}\) is a bounding-box represented as a tuple \((x_{ji}, y_{ji}, w_{ji}, h_{ji})\), where \((x_{ji}, y_{ji})\) denotes the coordinates of the top-left corner and \((w_{ji}, h_{ji})\) represent the width and height of the bounding-box, all in pixel coordinates relative to the corresponding image. The detection timestamp is set to the same value as the timestamp of the image in which the detection occurred. The inference process is optimized using tkDNN \cite{TKDNN}, a custom framework designed to enhance performance on NVIDIA GPUs using TensorRT technology. This optimization enables the model to achieve an average inference time of 4.5 ms for a batch of 7 input images on a NVIDIA RTX 6000 Ada GPU.

\subsection{LiDAR Detections}
\label{lidar_detections}
LiDAR object detection is performed using the PointPillars model \cite{POINTPILLARS}, which is initially pre-trained on the KITTI dataset \cite{KITTI} with the car label only. To adapt the model to specific race scenarios, a custom dataset has been created by manually labeling nearly 25000 LiDAR point clouds from previous race events.
During inference, the model receives a single point cloud composed of three LiDAR scans merged together with the respective extrinsic parameters and outputs a set of 3D detections (\(\textit{BB}^{\text{3D}}_{\text{lid}}\)). Each detection \(i\) is a 3D bounding-box represented as a tuple \((p_i, \theta_i, s_i)\), where \(p_i\) denotes the 3D position of the bounding-box within the point cloud, \(\theta_i\) represents its orientation relative to the z-axis, and \(s_i\) defines the dimensions of the box. Since LiDAR points are detected at different time instances due to the sensor's operating mode, the detection timestamp is calculated as the average of all contributing LiDAR points, which yields a more precise temporal reference for the tracker’s delay-compensation mechanism.
The inference process is optimized using tkDNN \cite{TKDNN}, enabling the model to achieve an average inference time of 8.4 ms on an NVIDIA RTX 6000 Ada GPU.

\subsection{RADAR Detections}
\label{RADAR_detections}
RADAR-based object detection is performed using a filtering and clustering approach. First, the RADAR point cloud is filtered by evaluating the RADAR Cross-Section (RCS), Signal-to-Noise Ratio (SNR), and the existence probability assigned to each detected point by the sensor. This step helps to eliminate outliers caused by reflections and ghosting effects. Next, using vehicle localization data, points outside the track boundaries are discarded. 
After filtering, a simple clustering algorithm is applied to the remaining points. Clusters are formed by imposing two constraints: the distance between all points within a cluster must not exceed a maximum distance \(\Delta d_{\text{max}}\), and the velocity difference between points must not be greater than \(\Delta v_{\text{max}}\). Only clusters with at least \(N_\text{min}\) points are considered valid. This entire processing pipeline is highly efficient, taking less than 3 ms for each RADAR point cloud.

\section{Detections Fusion}
\label{detections_fusion}
For each possible pair of detections from different sources with overlapping field-of-view, a matching process is carried out. The first step involves temporal matching, where pairs of detections are created based on their timestamps, ensuring that the time difference between them is less than $\Delta T_{\text{max}}$. Then the detection are merged based on their respective sources.

\subsection{Camera-LiDAR Matching}
\label{points_matching}
For each 2D bounding-box \(bb_{ji} \in \textit{BB}^{\text{2D}}_{\text{cam}_j}\), the LiDAR point clouds with overlapping field-of-view are selected. These point clouds are filtered by removing the ground plane. For each point, its normal vector is computed by fitting a local plane using its neighboring points. The points are then filtered based on the norm value along the vertical axis. Additionally, points above a certain height threshold (e.g., 3 meters) and points belonging to the ego-vehicle are discarded. The remaining LiDAR points are then projected onto the camera image using the transformation matrix \(\mathbf{T}^{\text{cam}_j}_{\text{lidar}_k}\), the intrinsic parameters of the camera \(\mathbf{K}_j\), and the distortion parameters \(\mathbf{d}_j\)
\[
    \mathbf{p}_{\text{cam}_j} = \text{distort}(\mathbf{K}_j \, \mathbf{T}^{\text{cam}_j}_{\text{lidar}_k} \, \mathbf{p}_{\text{lidar}_k}, \mathbf{d}_j)
\]
Only LiDAR points \(\textbf{P}_{\text{lidar}_k}\) whose corresponding image points \(\textbf{p}_{\text{cam}_j}\) fall inside the bounding-box \(bb_{ji}\) are considered. Finally, a distance histogram is constructed using the points range value, and the bin with the highest probability is selected to represent the object's distance. The final 3D detection is obtained by projecting the $bb_{ji}$ into world coordinates using the selected distance. The merged detection timestamp is recomputed as the mean timestamp of all contributing points.

\subsection{Camera-RADAR Matching}
\label{camera_RADAR_matching}
The same process used in the \textit{Camera-LiDAR Matching} is applied between each 2D bounding-box \(bb_{ji} \in \textit{BB}^{\text{2D}}_{\text{cam}_j}\) and each cluster detected in the RADAR point cloud, with the exception of the ground removal step. The merged detection includes additional information on the object's radial velocity obtained by the RADAR points. The detection timestamp is set to that of the RADAR point cloud, as it is the source that provides the distance information.

\subsection{LiDAR-RADAR Matching}
\label{lidar_RADAR_matching}
For each 3D bounding-box \(bb_i \in \textit{BB}^{\text{3D}}_{\text{lid}}\), the set of RADAR points within the bounding-box is identified. The merged detection is then created by augmenting the original \(bb_i\) with information on the object's radial velocity obtained by the RADAR points.

\section{Object Tracking}
\label{object_tracking}
The object detection pipeline described earlier provides observations that are subsequently used as input to a multi-object tracker. For each tracked object, its state is estimated through an Extended Kalman Filter (EKF).
Each object is also associated with a precomputed racing line that represents a trajectory it is most likely to follow along the track. The racing line encodes the position and heading of the reference path, as well as a speed profile describing the maximum achievable velocity of the race car at each point along the trajectory. Accordingly, the EKF state vector is expressed in Frenet coordinates relative to the associated racing line. The state is defined as $[s,d,v_s,v_d]$, where $s$ denotes the position of the object along the racing line, $d$ is the displacement of the object along the normal of the racing line, while $v_s$ and $v_d$ represent longitudinal and lateral velocities, respectively. 
The orientation of the object is inferred from the matched racing line using the curvilinear position $s$ of the object. This choice is motivated by the fact that PointPillars is the only detector that provides direct measurements of vehicle orientation. Consequently, in the presence of one or more LiDAR failures, alternative orientation corrections are not available.
The process model is derived from a modified constant-velocity formulation that introduces control inputs $[U_{v_s}, U_{v_d}]$ and constants $[\tau_{v_s}, \tau_{v_d}]$. This allows us to introduce first-order dynamics into the velocity-related state variables, as shown below
\begin{align}
    \dot{s} &= v_s \\
    \dot{d} &= v_d \\
    \dot{v}_s &= \frac{U_{v_s} - v_s}{\tau_{v_s}} \\
    \dot{v}_d &= \frac{U_{v_d} - v_d}{\tau_{v_d}}
\end{align}
Unlike a standard constant-velocity model, this formulation allows the predicted velocities to converge smoothly toward the reference values $[U_{v_s}, U_{v_d}]$. These references are derived from the speed profile associated with the selected racing line, thereby embedding prior knowledge of the expected vehicle behavior at different track segments into the prediction step. For example, the model naturally captures deceleration when approaching a corner and acceleration during corner exit, leading to more realistic and robust state predictions in high-speed racing scenarios. The constants $[\tau_{v_s}, \tau_{v_d}]$ define how fast the state converges to the reference values.

\subsection{Objects Life-cycle}
\label{tracklets_lifecycle}
The tracking module is responsible for updating active objects, discarding outdated ones, and creating new objects when necessary.
For each new detection, its timestamp is used to identify the two ego-vehicle states that immediately precede and follow it in time. The ego-vehicle state at the detection timestamp is then obtained via linear interpolation between these two states. The detection is then transformed into the global frame using the interpolated ego-vehicle pose.
Subsequently, if the detection timestamp is older than the current time, the delay compensation mechanism is applied to ensure the temporal consistency of the tracking state despite asynchronous and latency-affected detections. This is critical in high-speed racing scenarios, where even small timing mismatches can lead to meter-scale estimation errors. To implement this, the states of all tracked objects are reverted to the corresponding past instant using a maintained history of states and corrections for each object. A Munkres \cite{MUNKRES} association process is executed based on the Euclidean distance to associate the current detection with the reverted tracked objects. If the detection matches an existing object, it is inserted into its correction history at the right timing. After the association, all object states are reported to the current time, re-applying their correction history.
Otherwise, if no object matches the detection, a new object is created. The position is initialized with the detection position. The object velocity is initialized to the ego-vehicle velocity, as under normal conditions the two vehicles are expected to have similar speeds. In addition, the initial covariance is set high to describe the state uncertainty. The new object is considered tracked only after a minimum number of corrections (e.g. 3 to 5) are applied, ensuring robustness against sporadic false positives and stable convergence to its true state.
Active objects are removed based on the time since the last correction or if duplication is detected. More precisely, duplication occurs when the positions of two objects are too close to each other, in terms of squared Euclidean distance, considering their $N$ most recent states. In this case, the most recent object is discarded.

\subsection{Racing Line Matching}
\label{raceline_matching}
The tracking module relies on a set of precomputed racing lines, each associated with different speed profiles. These racing lines provide prior knowledge of conventional trajectories for the track. When a newly detected object is initialized, a default racing line is assigned to it, as a reference trajectory is required to express the object's state in Frenet coordinates.

As the object state evolves over time, a different racing line from the predefined set may better match the actual trajectory of the object and provide more accurate information about its heading and speed profile. Consequently, a racing line matching procedure is performed at each state update. Specifically, the cumulative Euclidean distance is computed between each candidate racing line and the last $N$ estimated states of the tracked object. The racing line that minimizes this cumulative distance is selected, and the object state is subsequently recomputed in the Frenet coordinate frame associated with the selected racing line.

\section{Experimental Results} \label{results}
During the 2025 A2RL event, multiple controlled test sessions were conducted with several autonomous race cars operating simultaneously on the track. On these occasions, sensor data has been collected onboard the race cars, and high-quality localization data for each vehicle was shared among test participants after the completion of each session. The localization data contains the position, heading, and longitudinal velocity of each vehicle on the track.
In this work, the shared localization data is used as ground truth, providing a reliable benchmark to assess the performance of the proposed tracking system. For evaluation purposes, both estimated and ground truth trajectories are transformed into Frenet coordinates with respect to the track center line. This representation enables the decomposition of the tracking error into longitudinal and lateral components. In this way, longitudinal error reflects discrepancies in the vehicle's position along the direction of motion, while lateral error represents deviations perpendicular to the track. This approach offers a more meaningful analysis of the tracking performance compared to traditional Euclidean error, as it takes into account the specific dynamics of the race cars. Evaluating the longitudinal error is particularly interesting in situations where the tracked object presents significant changes in speed due to hard braking or accelerating. The longitudinal velocity error and the orientation error are also computed to assess the accuracy of the estimation. 
To gain a more comprehensive understanding of the performance of the tracking system, the calculated errors are correlated with the distance between the ego-vehicle and the tracked object. This analysis enables the identification of patterns in the tracking error under various conditions on the track. In general, larger errors may be observed when the ego-vehicle is far from the opponent, highlighting situations where the system is less accurate. Analyzing how the tracking error varies under different race conditions helps to understand the strengths and limitations of the tracking pipeline and to identify specific scenarios where the detection and tracking system needs improvement or where its performance may be critical and requires accurate trajectory tracking.

This section demonstrates the necessity of a multi-modal pipeline by highlighting the complementary contributions of each modality to tracking accuracy through an ablation study. We further analyze the performance and limitations of the proposed method in critical racing scenarios that also reflect challenging urban edge cases. Specifically, we consider high relative speed overtaking, multi-object tracking under occlusions, and side-by-side overtaking.
\subsection{Ablation Study}
\begin{table}[b]
\caption{Sensor modalities ablation study results}
\resizebox{\linewidth}{!} {
\centering
\small
\setlength{\tabcolsep}{4pt}
\begin{tabular}{ccc|cccccc}
\toprule
 \multicolumn{3}{c|}{Modalities}
 & \multicolumn{1}{c}{Longitudinal}
 & \multicolumn{1}{c}{Lateral}
 & \multicolumn{1}{c}{Velocity} 
 & \multicolumn{1}{c}{Distance}
 & \multicolumn{1}{c}{Time}\\
 Camera & LiDAR & RADAR
 & RMSE [m] 
 & RMSE [m] 
 & RMSE [m/s] 
 & Max [m] 
 & [s] \\
\midrule

\xmark &\cmark &\cmark 
&\textbf{1.17}
&0.41
&0.95
&73.05
&308.6 
\\
\cmark &\xmark &\cmark 
&1.72
&0.53
&1.13
&95.15
&305.7
\\
\cmark &\cmark &\xmark 
&1.54
&0.42
&1.43
&76.91
&306.4
\\
\midrule
\cmark &\cmark &\cmark 
&1.40
&\textbf{0.41}
&\textbf{0.77}
&\textbf{96.0}
&\textbf{313.1}
\\
\bottomrule
\end{tabular}
}
\label{tab:ablation}
\end{table}
The ablation study is conducted on a 315-second sequence in which another vehicle is chased at high speed. Table~\ref{tab:ablation} reports the contribution of each sensing modality to the overall tracking performance and highlights how different sensors affect specific error components. The LiDAR-RADAR configuration achieves the lowest longitudinal RMSE, low lateral, and velocity errors, confirming the complementary nature of LiDAR spatial precision and RADAR Doppler information. When only camera and RADAR are used, performance degrades noticeably, with clear increases in both longitudinal and lateral errors. This highlights the critical role of LiDAR in providing accurate 3D detection, especially in the depth direction, where monocular camera estimates are inherently less accurate at long range. Using only camera and LiDAR yields intermediate performance. In this configuration, the velocity RMSE increases significantly, demonstrating the importance of RADAR Doppler measurements for accurate velocity estimation. Although the full multi-modal configuration exhibits a slightly higher longitudinal RMSE, it achieves the best overall balance across metrics. In particular, it provides the lowest velocity error, the highest lateral accuracy, and the longest tracking range and duration. The increase in longitudinal error is due to the extended tracking range: beyond the effective distance of LiDAR, only camera and RADAR detections are available. Even if these modalities are inherently less precise, they enable earlier object detection, improving overall tracking continuity. Overall, this ablation study highlights the complementary roles of each sensor: cameras enhance tracking continuity and range, RADAR improves motion estimation, and LiDAR provides precise 3D position information. These results validate the design of the proposed late-fusion architecture and confirm the importance of a unified multi-modal pipeline to fully exploit the heterogeneous information available from each sensing modality.

\subsection{High Relative Speed Difference Overtaking}
\begin{figure}[b]
\centering
{\includegraphics[width=\linewidth]{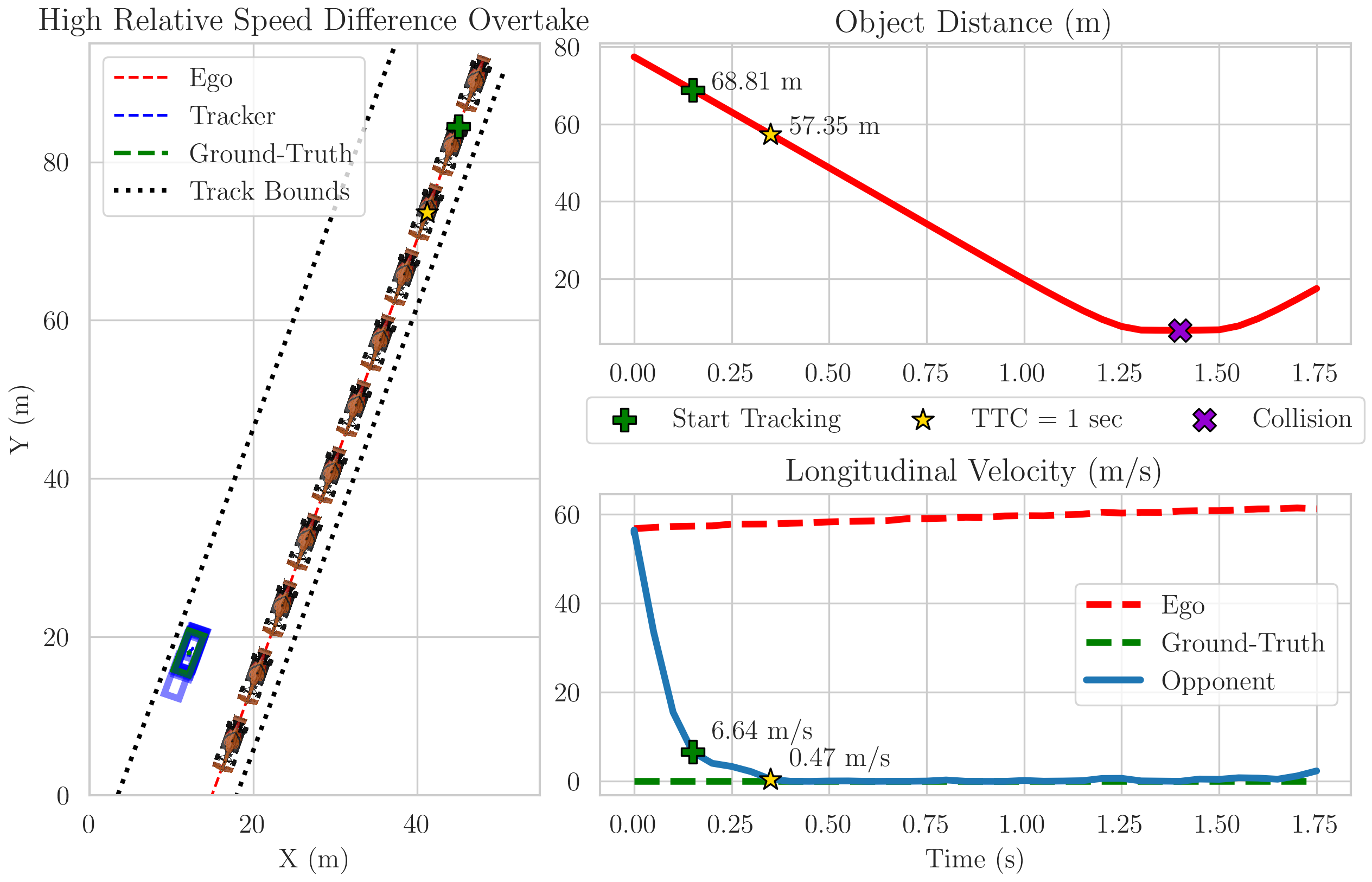}}
\caption{The object is initially detected at a distance of approximately 80 m and begins to be actively tracked once a sufficient number of detections are associated with it. At the start of tracking, the object is 68.81 m away, moving with an estimated velocity of 6.64 m/s. At the distance of 57.35 m, the TTC is reduced to 1 second, and the estimated velocity converged to nearly 0 m/s.}
\label{fig:high_delta_speed}
\end{figure}

A critical scenario in racing occurs when a vehicle suddenly stops on the track due to a technical failure, and another vehicle approaches it at high-speed from behind. This represents one of the worst-case situations, as the very high relative speed difference between the two vehicles drastically reduces the available reaction time. Consequently, it is essential to detect and begin tracking such objects as early as possible, while also obtaining a fast-converging and accurate velocity estimate. Reliable velocity estimation enables the planning module to accurately predict the time-to-collision with the stopped vehicle and to devise an appropriate evasive maneuver to avoid the collision.
Multiple controlled real-world tests were conducted with a stationary vehicle on the track. A specific case is presented in Fig. \ref{fig:high_delta_speed}. The stopped vehicle is first detected at a distance of approximately 80 m while the ego-vehicle is approaching from behind at nearly 60 m/s and accelerating. At this initial stage, a new object is created as it does not match any previously tracked object. Its velocity is initialized to the ego-vehicle speed. The object begins to be actively considered 150 ms after the first detection, once the minimum number of matched detections is reached. At this point, the object is 68.81 m from the ego-vehicle, with an estimated velocity of 6.64 m/s. Although the estimated speed of the object is not actually zero, the large relative speed difference between the ego-vehicle and the object provides critical information for the planning module to safely prepare the overtake maneuver. After 350 ms from the initial detection, the time-to-collision (TTC) reaches 1 s, representing the remaining time before a potential collision if no evasive action is taken. At this moment, the object is 57.35 m from the ego-vehicle, and its estimated velocity has converged to almost 0 m/s. The maximum tracking errors occur at the moment the object is first initialized, with a longitudinal error of 1.03 m, a lateral error of 0.25 m, and an orientation error of 0.90 degrees. After initialization, the estimate errors rapidly drop to almost zero, and the state remains stable.
This analysis demonstrates that the proposed tracking pipeline can detect stopped objects with high relative speed differences early, providing the planning module with sufficient time and accurate information to compute a safe and effective evasive trajectory.

\subsection{Multi-object Tracking Under Occlusions}
\begin{figure}[b]
\centering
{\includegraphics[width=\linewidth]{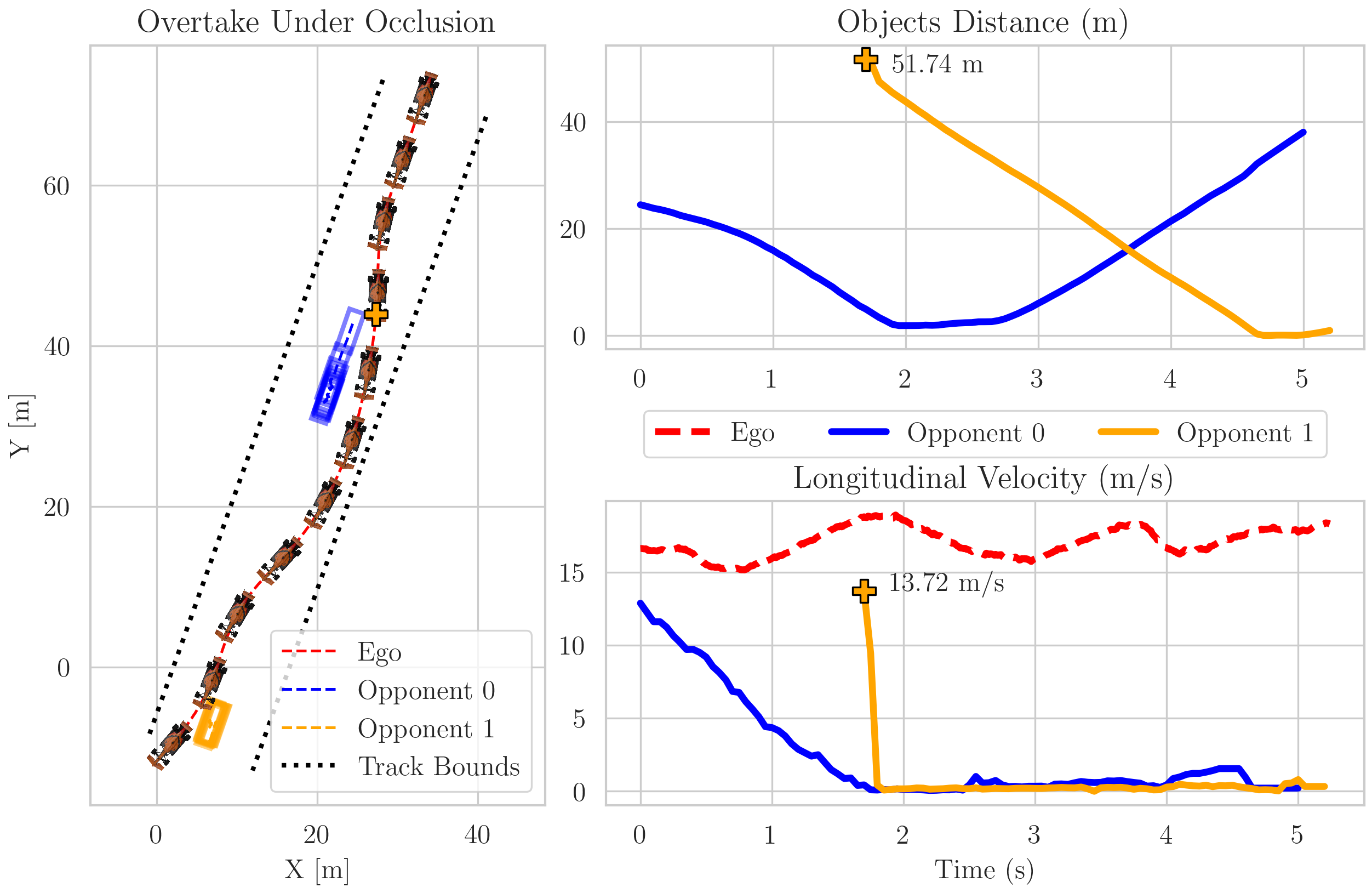}}
\caption{In this scenario, the \textcolor{blue}{blue} opponent occludes the visibility of the \textcolor{orange}{orange} opponent. As soon as the ego-vehicle initiates the overtaking maneuver, the \textcolor{orange}{orange} vehicle becomes visible at a distance of 51.74 m ahead, providing sufficient time for the planning module to adapt the overtaking trajectory.}
\label{fig:occlusion}
\end{figure}

In racing competitions, vehicles tend to follow similar trajectories to achieve optimal lap times. As a consequence, they are often longitudinally aligned, and the closest opponent is likely to occlude the visibility of other vehicles from the perspective of the ego-vehicle. In such conditions, when the ego-vehicle initiates an overtaking maneuver, a previously occluded opponent may suddenly appear in front of it. The detection and tracking pipeline must therefore minimize the latency required to initialize and stabilize the tracking of newly visible opponents in order to react safely.
In the scenario shown in Fig. \ref{fig:occlusion}, the ego-vehicle follows the \textcolor{blue}{blue} vehicle, which suddenly brakes with a deceleration of 7 m/s$^2$. In response, the planning module initiates an overtaking maneuver. As soon as the ego-vehicle moves laterally, a previously occluded and stopped vehicle (\textcolor{orange}{orange}) enters the field of view. This vehicle, which had not been tracked before, is first detected at a distance of approximately 52 m, and its estimated velocity rapidly converges to 0 m/s within the following two iterations. 
As in the previous scenario, the maximum tracking errors occur at the time of object initialization, with a longitudinal error of 0.29 m and a lateral error of 0.83 m. In contrast, the orientation error remains practically negligible throughout the tracking interval.
This scenario demonstrates that the proposed detection and tracking pipeline can immediately initialize the tracking for newly visible objects, providing sufficient time for the planning module to adapt the overtaking trajectory.

\subsection{Side-by-side Overtaking}
\begin{figure}[b]
\centering
{\includegraphics[width=\linewidth]{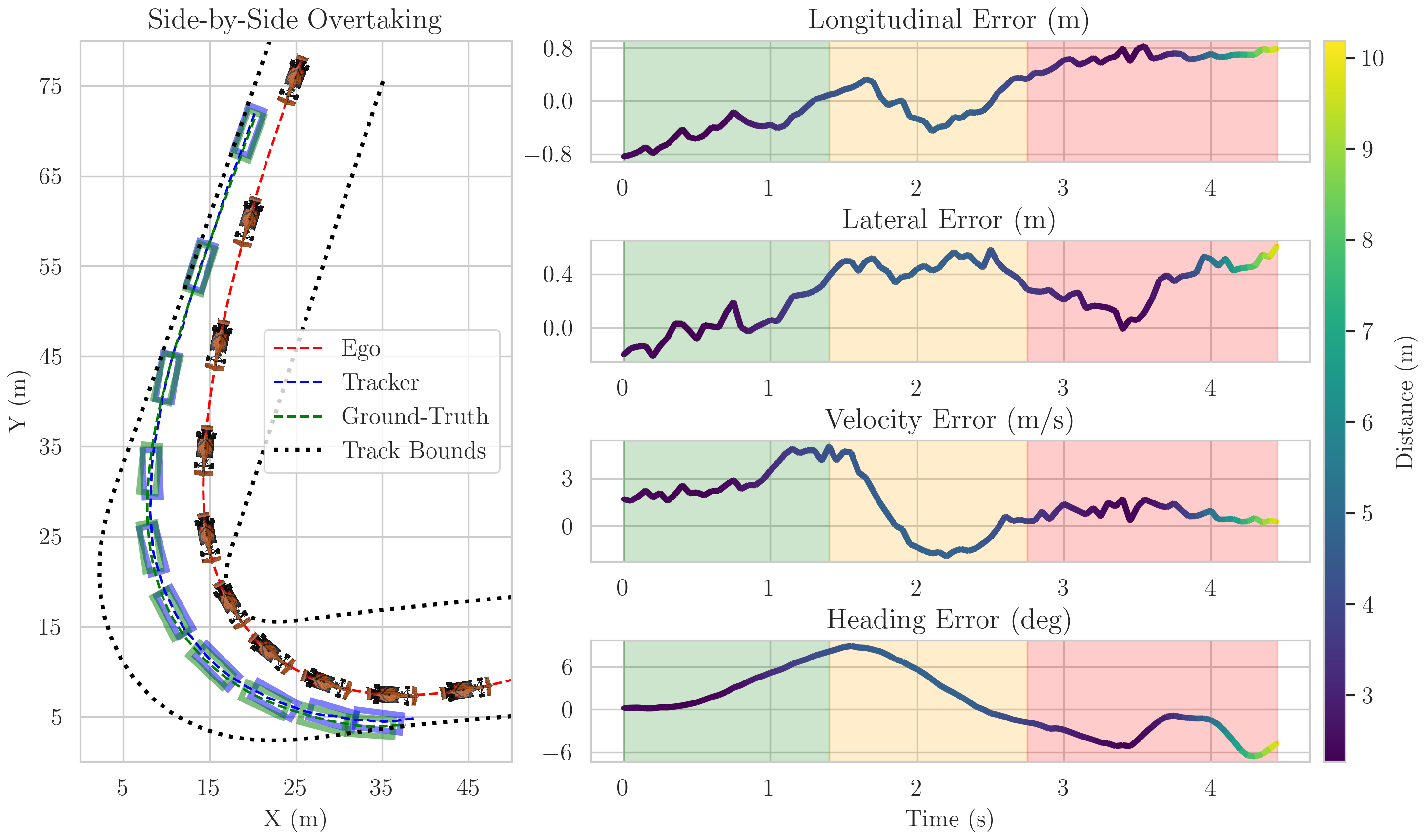}}
\caption{The side-by-side overtake maneuver is illustrated on the left, while the corresponding tracking errors are shown on the right. The error are correlated to the minimum distance between the two vehicles. The three colored sections indicate the relative position of the opponent vehicle with respect to the ego-vehicle: \textcolor{Green}{in front}, \textcolor{YellowOrange}{side-by-side}, and \textcolor{Red}{behind}, respectively.
}
\label{fig:side_by_side}
\end{figure}

Side-by-side overtaking represents one of the most critical situations in racing, as the vehicles operate at extremely close distances and near their handling limits. Ensuring the safety of both race cars and the successful execution of the overtaking maneuver requires high tracking accuracy. In this case study, we evaluate a side-by-side overtaking scenario in which our race car overtakes another vehicle, reaching a minimum lateral distance of 1.75 m. The analysis is shown in Fig. \ref{fig:side_by_side}. On the left, the ego-vehicle maneuver is shown in \textcolor{red}{red} and the opponent's estimated trajectory in \textcolor{blue}{blue} overlapped with the actual opponent ground-truth trajectory in \textcolor{Green}{green}. On the right, the tracking error with respect to the ground path is shown. The error line is colored based on the minimum distance between the ego-vehicle and the opponent ground-truth position. In addition, the three colored sections indicate the relative position of the opponent vehicle with
respect to the ego-vehicle: \textcolor{Green}{in front}, \textcolor{YellowOrange}{side-by-side}, and \textcolor{Red}{behind}, respectively.
The longitudinal exhibits a clear bias that depends on the relative position of the opponent vehicle. When the opponent is in front, the error is predominantly negative, it oscillates around zero when the vehicles are laterally aligned, while it increases during the overtaking phase and remains positive once the opponent is consistently behind the ego-vehicle. This behavior is mainly attributed to the tracker correction being dominated by the fusion of camera detections with the other sensors. Our camera-based detections lack semantic awareness of which specific part of the opponent vehicle is actually being observed. As a result, when the opponent is ahead, the tracker mainly observes the rear of the vehicle and underestimates the longitudinal position. Conversely, when the opponent is behind, the tracker mainly observes the front part of the vehicle, leading to an overestimation of the longitudinal position. The lateral error is also affected by a similar bias, but only when the vehicles are laterally aligned. This bias results in a maximum longitudinal error up to 0.84 m and a maximum lateral error up to 0.61 m. Additionally, the lateral error observed when the opponent vehicle is behind is influenced by the reduced number of measurement updates. In this configuration, rear visibility is provided only by the rear camera and the rear RADAR, as the LiDARs are occluded by the rear wing. The estimated longitudinal velocity initially exhibits an overshoot, reaching a peak error of 5 m/s, and fails to fully capture the opponent’s abrupt braking maneuver. It then gradually converges to the correct value in approximately 1.75 seconds. Finally, the heading error is largely driven by the racing line selected from the set provided to the tracker. The chosen racing line assumes a tighter cornering trajectory, whereas the opponent vehicle instead reacted to the ego-vehicle taking a wider turn. This mismatch results in a positive increase in heading error, up to 9 degrees. As following the taken trajectory would have led the opponent outside the track boundaries, the opponent steered left earlier than expected, causing the heading error to shift toward negative values.
This analysis clearly highlights that estimation accuracy strongly depends on the relative position of the tracked object with respect to the ego-vehicle, as well as on the need to incorporate a more complete semantic understanding of vehicle geometry into the detection process. Despite the limitations discussed, the achieved estimation accuracy was sufficient to successfully complete the side-by-side overtaking maneuver.

\section{CONCLUSION}
\label{conclusion}
The proposed architecture shows that combining multi-modal sensing with delay-aware tracking and dynamics-informed priors is key for reliable perception in extreme-speed environments. A late-fusion scheme improves the robustness to sensor failures, while the tracker compensates for detection delays and incorporates prior knowledge of the vehicle and track. Real-world racing experiments validate the effectiveness of the pipeline for safe high-speed planning. Future work will address the observed limitations by incorporating semantic-aware vehicle geometry models, enabling the tracker to explicitly reason about which portion of the opponent vehicle is being observed and reducing estimation biases.



\bibliographystyle{IEEEtran}
\bibliography{refs} 

@inproceedings{Kulkarni_2023,
  author = {Kulkarni \textit{et al.}, A.},
  title = {RACECAR - The Dataset for High-Speed Autonomous Racing},
  booktitle = {Proc. 2023 IEEE/RSJ Int. Conf. Intell. Robots Syst. (IROS)},
  year = {2023},
  month = {Oct.},
  url = {http://dx.doi.org/10.1109/IROS55552.2023.10342053}
}

@inproceedings{POINTPILLARS,
  author = {Lang, A. H. and Vora, S. and Caesar, H. and Zhou, L. and Yang, J. and Beijbom, O.},
  title = {PointPillars: Fast Encoders for Object Detection From Point Clouds},
  booktitle = {Proc. 2019 IEEE/CVF Conf. Comput. Vis. Pattern Recognit. (CVPR)},
  year = {2019},
  pages = {12689--12697},
  doi = {10.1109/CVPR.2019.01298}
}

@article{Jung_2023,
  author = {Jung \textit{et al.}, C.},
  title = {An autonomous racing system: Design, implementation, and analysis; team KAIST at the IAC},
  journal = {Field Robotics},
  volume = {3},
  pages = {766--800},
  year = {2023}
}

@article{cellina2025,
  author = {Cellina, M. and Corno, M. and Savaresi, S. M.},
  title = {LiDAR-Based Vehicle Detection and Tracking for Autonomous Racing},
  journal = {arXiv preprint},
  volume = {arXiv:2501.14502},
  year = {2025},
  url = {https://arxiv.org/abs/2501.14502}
}

@article{betz,
  author = {Betz, J. and Zheng, H. and Liniger, A. and Rosolia, U. and Karle, P. and Behl, M. and Krovi, V. and Mangharam, R.},
  title = {Autonomous Vehicles on the Edge: A Survey on Autonomous Vehicle Racing},
  journal = {IEEE Open J. Intell. Transp. Syst.},
  volume = {3},
  pages = {458--488},
  year = {2022},
  doi = {10.1109/OJITS.2022.3181510}
}

@article{Mao_2023,
  author = {Mao, J. and Shi, S. and Wang, X. and others},
  title = {3D Object Detection for Autonomous Driving: A Comprehensive Survey},
  journal = {Int. J. Comput. Vis.},
  volume = {131},
  pages = {1909--1963},
  year = {2023},
  doi = {10.1007/s11263-023-01790-1}
}

@article{YOLOV4,
  author       = {Alexey Bochkovskiy and
                  Chien{-}Yao Wang and
                  Hong{-}Yuan Mark Liao},
  title        = {YOLOv4: Optimal Speed and Accuracy of Object Detection},
  journal      = {CoRR},
  volume       = {abs/2004.10934},
  year         = {2020},
  url          = {https://arxiv.org/abs/2004.10934},
  eprinttype    = {arXiv},
  eprint       = {2004.10934},
  bibsource    = {dblp computer science bibliography, https://dblp.org}
}

@article{BBD100K,
  author       = {Fisher Yu and
                  Wenqi Xian and
                  Yingying Chen and
                  Fangchen Liu and
                  Mike Liao and
                  Vashisht Madhavan and
                  Trevor Darrell},
  title        = {{BDD100K:} {A} Diverse Driving Video Database with Scalable Annotation
                  Tooling},
  journal      = {CoRR},
  volume       = {abs/1805.04687},
  year         = {2018},
  url          = {http://arxiv.org/abs/1805.04687},
  eprinttype    = {arXiv},
  eprint       = {1805.04687},
  bibsource    = {dblp computer science bibliography, https://dblp.org}
}

@inproceedings{TKDNN,
  title={A Systematic Assessment of Embedded Neural Networks for Object Detection},
  author={Verucchi \textit{et al.}, Micaela},
  booktitle={2020 25th IEEE International Conference on Emerging Technologies and Factory Automation (ETFA)},
  volume={1},
  pages={937--944},
  year={2020},
  organization={IEEE}
}

@article{KITTI,
  author = {Andreas Geiger and Philip Lenz and Christoph Stiller and Raquel Urtasun},
  title = {Vision meets Robotics: The KITTI Dataset},
  journal = {International Journal of Robotics Research (IJRR)},
  year = {2013}
}

@INPROCEEDINGS{PTP,

  author={Watt, Steve T. and Achanta, Shankar and Abubakari, Hamza and Sagen, Eric and Korkmaz, Zafer and Ahmed, Husam},

  booktitle={2015 Saudi Arabia Smart Grid (SASG)}, 

  title={Understanding and applying precision time protocol}, 

  year={2015},

  volume={},

  number={},

  pages={1-7},

  doi={10.1109/SASG.2015.7449285}}

@misc{ghosh2025racingdatasetbaselinemodel,
      title={A Racing Dataset and Baseline Model for Track Detection in Autonomous Racing}, 
      author={Shreya Ghosh and Yi-Huan Chen and Ching-Hsiang Huang and Abu Shafin Mohammad Mahdee Jameel and Chien Chou Ho and Aly El Gamal and Samuel Labi},
      year={2025},
      eprint={2502.14068},
      archivePrefix={arXiv},
      primaryClass={cs.CV},
      url={https://arxiv.org/abs/2502.14068}, 
}

@article{deng2020voxel,
  title={Voxel R-CNN: Towards High Performance Voxel-based 3D Object Detection},
  author={Deng, Jiajun and Shi, Shaoshuai and Li, Peiwei and Zhou, Wengang and Zhang, Yanyong and Li, Houqiang},
  journal={arXiv:2012.15712},
  year={2020}
}

@INPROCEEDINGS{RADAR_CALIBRATION,
  author={Doer, Christopher and Trommer, Gert F.},
  booktitle={2020 European Navigation Conference (ENC)}, 
  title={Radar Inertial Odometry With Online Calibration}, 
  year={2020},
  volume={},
  number={},
  pages={1-10},
  doi={10.23919/ENC48637.2020.9317343}}

@INPROCEEDINGS{CAMERA_LIDAR_CALIBRATION,
  author={Chen, Shaohui and Zhang, Huiquan and Yang, Zixuan and Yang, Yahui and Wang, Kewei},
  booktitle={2024 IEEE 25th China Conference on System Simulation Technology and its Application (CCSSTA)}, 
  title={ArUco Code-Based Calibration Method for LiDAR and Camera Integration}, 
  year={2024},
  volume={},
  number={},
  pages={674-678},
  doi={10.1109/CCSSTA62096.2024.10691812}}

@techreport{CAMERA_CALIBRATION,
author = {Burger, Wilhelm},title = {Zhang’s Camera Calibration Algorithm: In-Depth Tutorialand Implementation},language = {english},institution = {University of Applied Sciences Upper Austria, School ofInformatics, Communications and Media, Dept. of DigitalMedia},address = {Hagenberg, Austria},number = {HGB16-05},year = {2016},month = {05},url = {https://www.researchgate.net/publication/303233579_Zhang’s_Camera_Calibration_Algorithm_In-Depth_Tutorial_and_Implementation}}

@INPROCEEDINGS{BETTY_DATASET,
  author={Nye, Micah and Raji, Ayoub and Saba, Andrew and Erlich, Eidan and Exley, Robert and Goyal, Aragya and Matros, Alexander and Misra, Ritesh and Sivaprakasam, Matthew and Bertogna, Marko and Ramanan, Deva and Scherer, Sebastian},
  booktitle={2025 IEEE International Conference on Robotics and Automation (ICRA)}, 
  title={BETTY Dataset: A Multi-Modal Dataset for Full-Stack Autonomy}, 
  year={2025},
  volume={},
  number={},
  pages={2453-2460},
  doi={10.1109/ICRA55743.2025.11127350}}

@ARTICLE{MIT_PIT_PERCEPTION,
  author={Saba \textit{et al.}, Andrew},
  journal={Field Robotics}, 
  title={Fast and Modular Autonomy Software for Autonomous Racing Vehicles}, 
  year={2024},
  volume={4},
  number={},
  pages={1-45},
  doi={10.55417/fr.2024001}}

@article{AB3DMOT, 
author = {Weng, Xinshuo and Wang, Jianren and Held, David and Kitani, Kris}, 
journal = {IROS}, 
title = {{3D Multi-Object Tracking: A Baseline and New Evaluation Metrics}}, 
year = {2020} 
}

@ARTICLE{TUM_PERCEPTION,
  author={Karle, Phillip and Fent, Felix and Huch, Sebastian and Sauerbeck, Florian and Lienkamp, Markus},
  journal={IEEE Transactions on Intelligent Vehicles}, 
  title={Multi-Modal Sensor Fusion and Object Tracking for Autonomous Racing}, 
  year={2023},
  volume={8},
  number={7},
  pages={3871-3883},
  doi={10.1109/TIV.2023.3271624}}

@INPROCEEDINGS{OT_LOGIC,
  author={Toschi, Alessandro and Prignoli, Francesco and Bertogna, Marko},
  booktitle={2025 IEEE/RSJ International Conference on Intelligent Robots and Systems (IROS)}, 
  title={Modular Decision-Making and Drivable Areas for Multi-Agent Autonomous Racing}, 
  year={2025},
  volume={},
  number={},
  pages={12435-12441},
  doi={10.1109/IROS60139.2025.11246897}}

@article{MUNKRES,
author = {Munkres, James},
title = {Algorithms for the Assignment and Transportation Problems},
journal = {Journal of the Society for Industrial and Applied Mathematics},
volume = {5},
number = {1},
pages = {32-38},
year = {1957},
doi = {10.1137/0105003},
URL = { 
        https://doi.org/10.1137/0105003 
},
eprint = { 
        https://doi.org/10.1137/0105003
}
}

@misc{KAIST_MONODINODETR,
      title={MonoDINO-DETR: Depth-Enhanced Monocular 3D Object Detection Using a Vision Foundation Model}, 
      author={Jihyeok Kim and Seongwoo Moon and Sungwon Nah and David Hyunchul Shim},
      year={2025},
      eprint={2502.00315},
      archivePrefix={arXiv},
      primaryClass={cs.CV},
      url={https://arxiv.org/abs/2502.00315}, 
}

\end{document}